\documentclass[letterpaper, 10 pt, conference]{ieeeconf}  

\usepackage{graphicx}
\usepackage{bm}
\usepackage{caption}
\usepackage{subcaption}

\IEEEoverridecommandlockouts                              

\usepackage{graphics} 
\usepackage{epsfig} 
\usepackage{mathptmx} 
\usepackage{times} 
\usepackage{amsmath} 
\usepackage{amssymb}  

\title{\LARGE \bf
Aerial Mobile Manipulator System to Enable Dexterous Manipulations with Increased Precision
}

\author{Abbaraju Praveen$^{1}$, Haoguang Yang$^{1}$, Hyukjun Jang$^{2}$ and Richard M Voyles$^{3}$
\thanks{$^{1}$PhD Student at School of Engineering Technology, Purdue Polytechnic Institute, Purdue University, 47907, IN, USA.
        {\tt\small \{abbaraju,\ yang1510\}@purdue.edu}}%
\thanks{$^{2}$Undergraduate student at Mechanical Engineering, Purdue University, 47907, IN, USA.
        {\tt\small jang66@purdue.edu}}%
\thanks{$^{3}$Faculty at School of Engineering Technology, Purdue Polytechnic Institute, Purdue University, 47907, IN, USA.
        {\tt\small rvoyles@purdue.edu}}%
}

\begin{document}

\maketitle
\thispagestyle{empty}
\pagestyle{empty}

\begin{abstract}

Problems associated with physical interactions using aerial mobile manipulators (AMM) are being independently addressed with respect to mobility and manipulability. Multirotor unmanned aerial vehicles (UAV) are a common choice for mobility while on-board manipulators are increasingly be used for manipulability. However, the dynamic coordination between the UAV and on-board manipulator remains a significant obstacle to enable dexterous manipulation with high precision. This paper presents an AMM system configuration to addresses both the mobility and manipulability issues together. A fully-actuated UAV is chosen to achieve dexterous aerial mobile manipulation, but is limited by the actuation range of the UAV. An on-board manipulator is employed to enhance the performance in terms of dexterity and precision at the end-effector. Experimental results on position keeping of the dexterous hexrotor by withstanding the disturbances caused by the motions of the on-board manipulator and external wind disturbances are presented. Preliminary simulation results on end-point tracking in a simple planar on-board manipulator case is presented.

\end{abstract}

\section{INTRODUCTION}

Aerial mobile manipulation has been showing promising results in providing both high mobility and manipulability. Aerial mobility is achieved using multirotor UAVs, for their ability to hover, navigate and perform aggressive maneuvers in confined spaces, compared to fixed-wing UAVs. The manipulability is achieved by employing on-board manipulator mechanism of at least 1 degree of freedom (DOF).  Researchers have considered various designs of aerial mobile manipulator (AMM) systems to enable physical interaction with the environment. Some of the works targeted towards physical interaction \cite{c1} are aerial grasping \cite{c2} and transportation \cite{c3} \cite{c5}, object manipulation \cite{c4}, contact-based inspection \cite{c6} [14], etc.

Physical interactions are primarily achieved by just the UAV equipped with a passive 1-DOF manipulator on-board \cite{c6} \cite{c10} \cite{c15}. This type of AMM system usually has a custom end-effector suitable for the applications, such as gasping an object, physical contact, etc. However, these AMM systems lack dexterity at the end-effector, as the commonly employed UAVs such as quadrotors are inherently under-actuated \cite{c7} \cite{c8}. A fully-actuated or an over-actuated UAVs \cite{c16} is better suited to achieving high-precision in physical interaction. 

For high precision and high fidelity tasks, UAV mobile base is required to withstand the disturbances and provide a steady base for physical interactions. The Dexterous Hexrotor is a type of fully-actuated UAV which can exert an arbitrary wrench and provide a precise mobile base \cite{c9} for AMM system. With tilted rotors in \textit{cant} and \textit{dihedral} angles, this UAV can achieve precision mobility and manipulability within its actuation limits and configuration. However, with an additional DOF manipulator on-board, dexterity at the end-effector can be extended to a greater range.

This paper presents an AMM system in macro/ micro combination with Dexterous Hexrotor UAV as macro and a 6-DOF parallel manipulator as on-board micro manipulator. The macro/micro design results in optimal dynamic performance with reduced inertial properties at the end-effector \cite{c11}. In general, parallel manipulators have higher precision with reduced inertia at the end-effector, making it best choice for the AMM system. However, the parallel manipulator movements causes internal disturbances within the AMM system which can be rejected by coupling both the systems using operational space (OP) formulations \cite{c16}. The OP provides the decomposition of the forces into the forces at end-effector and forces required to reject the internal disturbances. This is presented in detail in section II. 

  \begin{figure}[th]
      \centering 
        \framebox{
        \includegraphics[width = 0.30\textwidth]{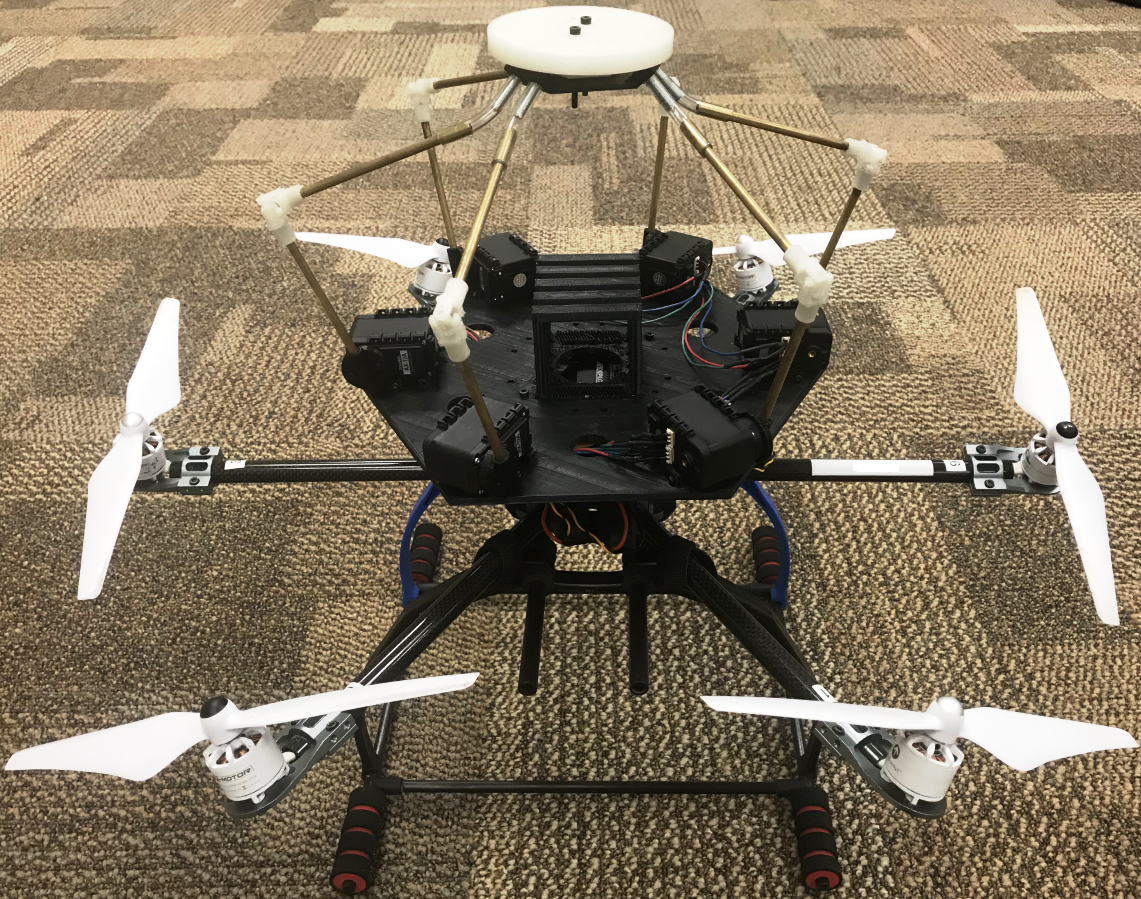}
}
      \caption{Prototype design of an AMM system with parallel manipulator facing upwards.}
      \label{AMM}
  \end{figure}

A unique configuration the AMM system with the parallel manipulator placed on top of the UAV by facing upwards as shown in fig. \ref{AMM} is discussed in the paper. This configuration of AMM system allows it to interact with the surfaces of high-rising structures such as high bridges or sky-scrapers, where ground equipment is unable to reach, and aloft work may pose a challenge to physical and mental safety of personnel. This paper is aimed to address the problems associated with enabling dexterous physical interactions with high precision manipulations. Preliminary results in enabling the UAV to withstand the motions of the on-board parallel manipulator are presented in the section IV. We have also presented the simulation results of end-point tracking for the AMM system with dexterous hexrotor and 2-DOF on-board planar manipulator.

\section{AERIAL MOBILE MANIPULATOR SYSTEM}

An AMM system is designed in macro-micro combination, with UAV as macro for high mobility and an on-board manipulator as micro for higher precision. The macro manipulator is designed to stabilize against all motions of micro manipulator. The micro manipulator is designed with reduced inertial properties compared to the macro manipulator and should be able to stabilize itself against all macro motions. The AMM system in macro/ micro combination results in optimal dynamic performance with reduced inertial properties at the end-effector \cite{c11} yielding in higher performance for force exertion. 

\subsection{UAV- Dexterous Hexrotor}

\begin{figure}[thpb]
      \centering
      \framebox{
      \includegraphics[width = 0.3\textwidth]{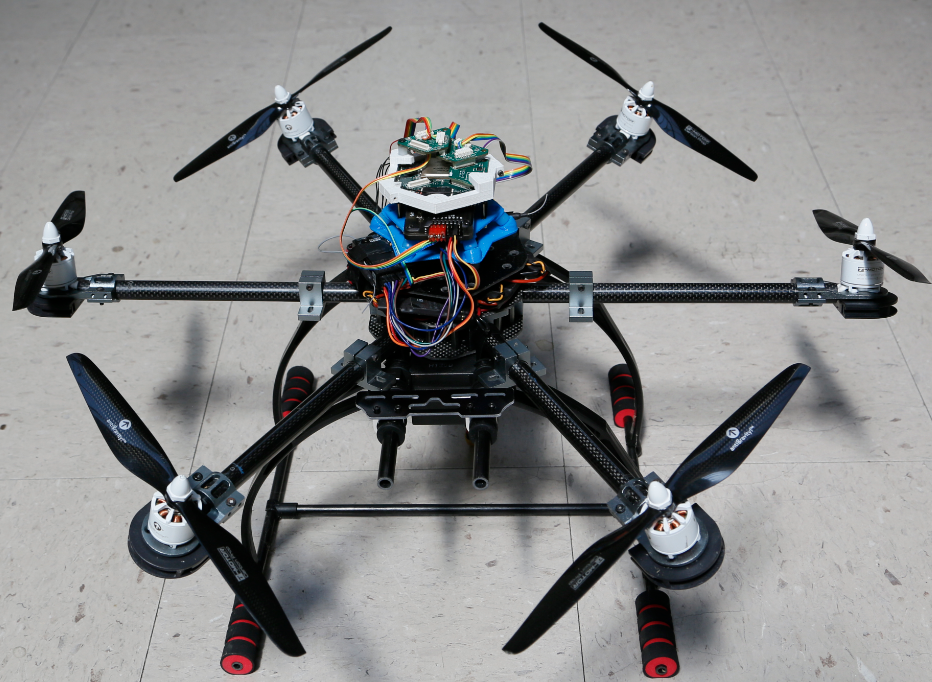}
} 
      \caption{Dexterous Hexrotor UAV prototype with $28^\circ$ cant angle and $0^\circ$ dihedral angle. 
      \label{Dexhex}}
      
   \end{figure}

The Dexterous Hexrotor is a class of full-actuated UAV, which is designed from hexrotor platform, with non-parallel rotor configuration. The non-parallel configuration is achieved by rotating each rotor at predefined angles, \textit{cant} angle, $\alpha$ and \textit{dihedral} angle, $\beta$, to achieve full actuation in 6-DOF as shown in fig.~\ref{Dexhex}. The \textit{cant} angle is achieve by rotating the adjacent rotors in opposite direction. Whereas, the \textit{dihedral} angle is achieved by rotating the rotors inwards/ outwards. As a result, the mapping matrix, $M_{(\alpha, \beta)}$ in (1) which maps the 6-D actuator space ($\omega^2$) to 6-D force/torque space ($f$ - forces; $\tau$ - torques at the end-effector), has full rank as shown in (2). Full actuation is possible by mapping the rotor thrusts to 6D Cartesian force space.

\begin{align}
\begin{bmatrix}
\bm{f}\\
\bm{\tau}
\end{bmatrix}
=
M_{(\alpha,\beta)}
\cdot
\bm{\omega^2}
\\
\textnormal{rank}\left(\bm{M_{(\alpha,\beta)}}\right)
= 6
\end{align}

The authors in \cite{c13} presented an optimization procedure to obtain optimal \textit{Cant} angle and \textit{dihedral} angle with respect to the efficiency and horizontal force exertion. The cant angle was suggested to be $28^\circ$ for higher horizontal force exertion, required for physical interaction and $18^\circ$ to yield higher efficiency and holonomic motion.

\subsection{On-board Manipulator}

The on-board manipulator design should be able to stabilize itself against all motions of the UAV (macro) \cite{c11}. This can be realized by choosing a manipulator with reduced inertial properties at the end-effector, compared that of UAV. A parallel manipulator design is chosen for its lower inertia and higher forces at the end-effector, compared to serial chain manipulators. 

\begin{figure}[thpb]
      \centering
      \framebox{
      \includegraphics[scale=0.2]{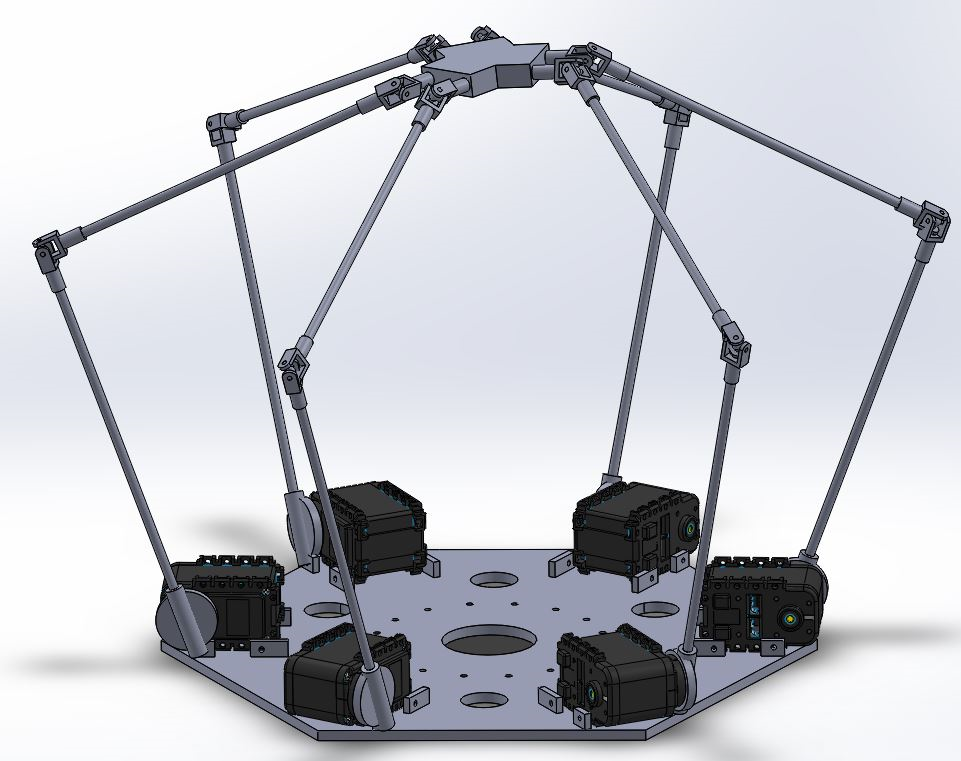}
}
      \caption{Light-weight 6RUS Parallel manipulator design.}
      \label{PM}
   \end{figure}
   
A parallel manipulator with 6-DOF would enhance the precision and dexterity at the end-effector of AMM system. However, the downside of such designs are high degrees of design complexity and higher stiffness, which are solvable using coupled dynamics and impedance control scheme. A basic hexa design with 6 RUS (revolute, universal \& spherical) parallel manipulator, as shown in fig. \ref{PM}, is employed here. It is designed to have light weight structure with most of its weight concentrated on the base of the manipulator. When attached to the UAV as a part of AMM system, this design will ensure the COM of the system doesn't shift far away from the COM of the UAV.

\begin{align}
\bm{\Gamma}
&=
\bm{J_{pm}^T \ F}
\\
\bm{J_{pm}}
&=
\bm{J_\gamma^{-1} \ J_X}
\end{align}
where, $\Gamma$ is joint torques vector, $F$ is the end-effector force vector, $J_{pm}$ is the Jacobian of the parallel manipulator, $J_{\gamma}$ is the Jacobian in the joint space and $J_{X}$ is the Jacobian in the end-effector space.

\subsection{Force Decomposition}

The force decomposition of the redundant AMM with 12 DOF is performed by employing the OS formulations. The dynamically consistent relationship between the operational forces, $F_{op}$, with the joint torques and rotor speeds of the AMM system is given in (\ref{force}).

\begin{eqnarray}
\begin{bmatrix}
\bm{\Gamma}\\
\bm{\omega^2}
\end{bmatrix}
=
\bm{J^T \ F_{op}}
+
\bm{[I-J^T \bar{J}^T] \ \Gamma_0}
\label{force}
\end{eqnarray}
where,
\begin{align}
\bm{F_{op}}
&=
\bm{\Lambda(q) \ {F_m}^*}
\\
\bm{J}
&=
\begin{bmatrix}
\bm{J_{pm}}\\
\bm{M_{(\alpha,\beta)}}
\end{bmatrix}
\end{align}

This decomposition of joint torques and rotor speeds into decoupled control vectors, as force at end-effector and joint motions in nullspace. The joint torques and rotor speeds corresponding to the forces at the operational point is given as, $J^T F_{op}$. This generalized operational force vector derived by mapping the desired force, using mass matrix $\Lambda(q)$. is  Whereas, the joint torques and rotor speeds rejecting the internal disturbances is provided by $[I-J^T \bar{J}^T]\Gamma_0$.

\section{PRELIMINARY EXPERIMENTAL RESULTS}

In achieving dexterous physical interaction with the environment, the AMM system should be able to withstand the disturbances caused by the coupled motion and maintain the end-effector position. The disturbances usually caused by the on-board manipulator motions, physical impact, wind turbulence while flying close to structures. The proposed work still needs development in-terms truly exerting coupled motion. Therefore, we are presenting preliminary results on the disturbance rejection of the UAV for all micro manipulator motions.

\begin{figure}[htb]
      \centering
      \framebox{
      \includegraphics[scale=0.7]{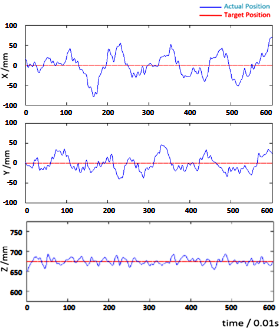}
      }
      \caption{Position keeping with respect to XYZ-axes under the disturbances caused by the motion of the on-board parallel manipulator.}
      \label{response}
   \end{figure}

The experiments are conducted with the AMM system flying in position control under motion capture system (VICON). A fine tuned nested P-PD controller is employed for the UAVs during the experiments. The parallel manipulator is tele-operated while the UAV is holding its position. The disturbances caused by the parallel manipulator motions on the UAV are plotted in fig. \ref{response}. 

The Dexterous Hexrotor has higher control bandwidth in rejecting the disturbances compared to conventional multirotor UAVs such as quadrotors. We conducted experiments to show the effectiveness of Dexterous Hexrotor in holding its position under wind disturbances. The goal of the experiment is to measure the position of the UAV when under free hover and under wind disturbances. Table I, summarizes the measurements as the standard deviation of the error. The reduced percentage of error in position keeping, under the presence of external disturbances, shows that higher precision with respect to the end-effector can be achieved. 

\begin{table}[h]
\caption{Standard deviation of position error in x-axis of Dexterous Hexrotor \& quadrotor during each period and increased error under wind disturbances.}
\label{table_example}
\begin{center}
\begin{tabular}{|c||c|c|}
\hline
Configuration  & Quadrotor & Dexterous Hexrotor\\
\hline
\hline
Free Hovering & 27.0054 mm & 25.3689 mm\\
\hline
Under Disturbance & 47.7000 mm & 41.2959 mm\\
\hline
Increased Error & 76.63\%  & 62.78\% \\
\hline
\end{tabular}
\end{center}
\end{table}

\subsection{End-point tracking with 2DOF manipulator}

\begin{figure}[th]
      \centering 
        \framebox{
        \includegraphics[width = 0.35\textwidth]{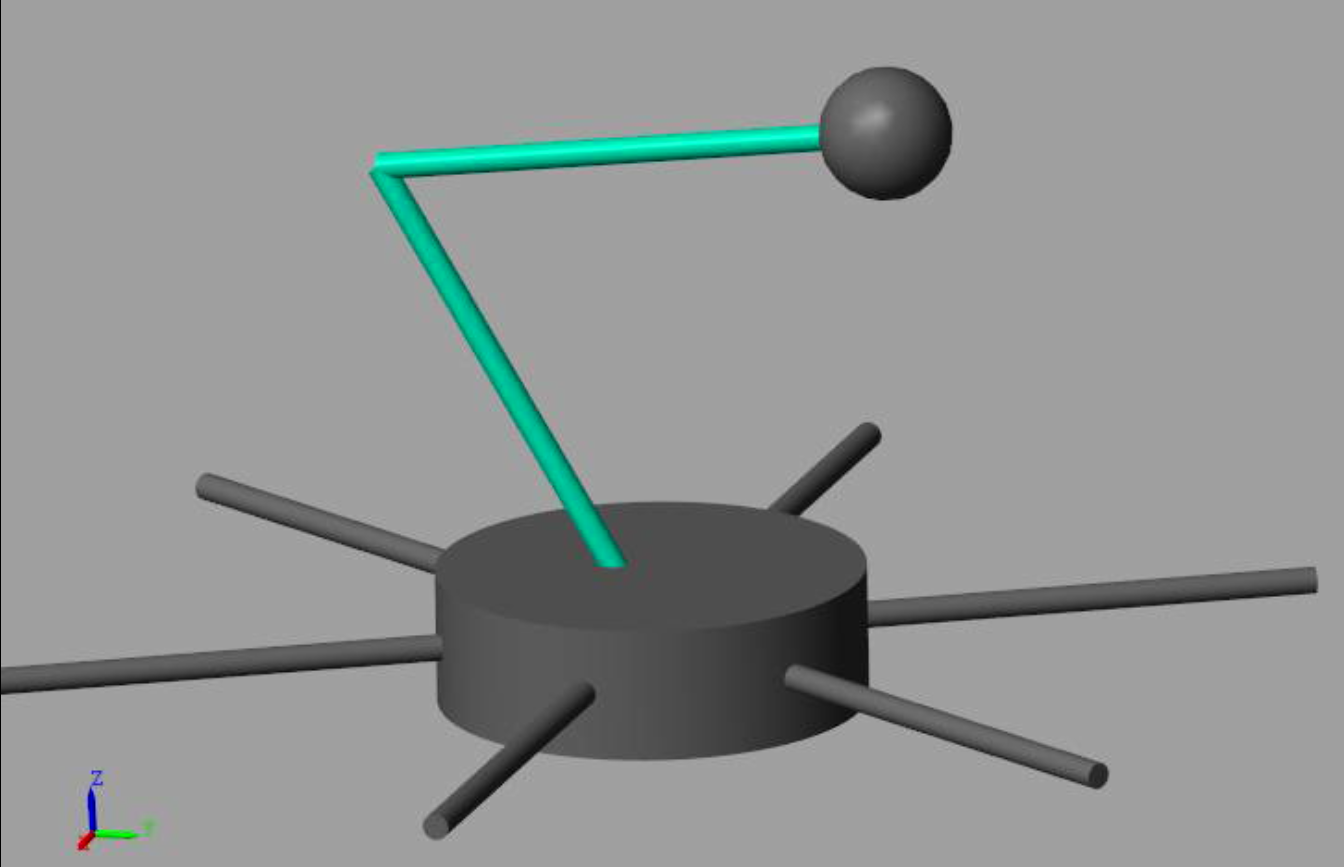}
}
      \caption{AMM system with 2-DOF planar manipulator facing upwards.}
      \label{AMM2}
  \end{figure}

The end-point tracking of the AMM system under wind disturbances shows the effectiveness of the coupled motion of the UAV and the manipulator. However, as a part of preliminary results, we are presenting the simulation results for a simple planar case, where the Dexterous Hexrotor is equipped with a 2-DOF planar manipulator as shown in fig.\ref{AMM2}.

\begin{figure}[th]
      \centering 
        \framebox{
        \includegraphics[width = 0.46\textwidth]{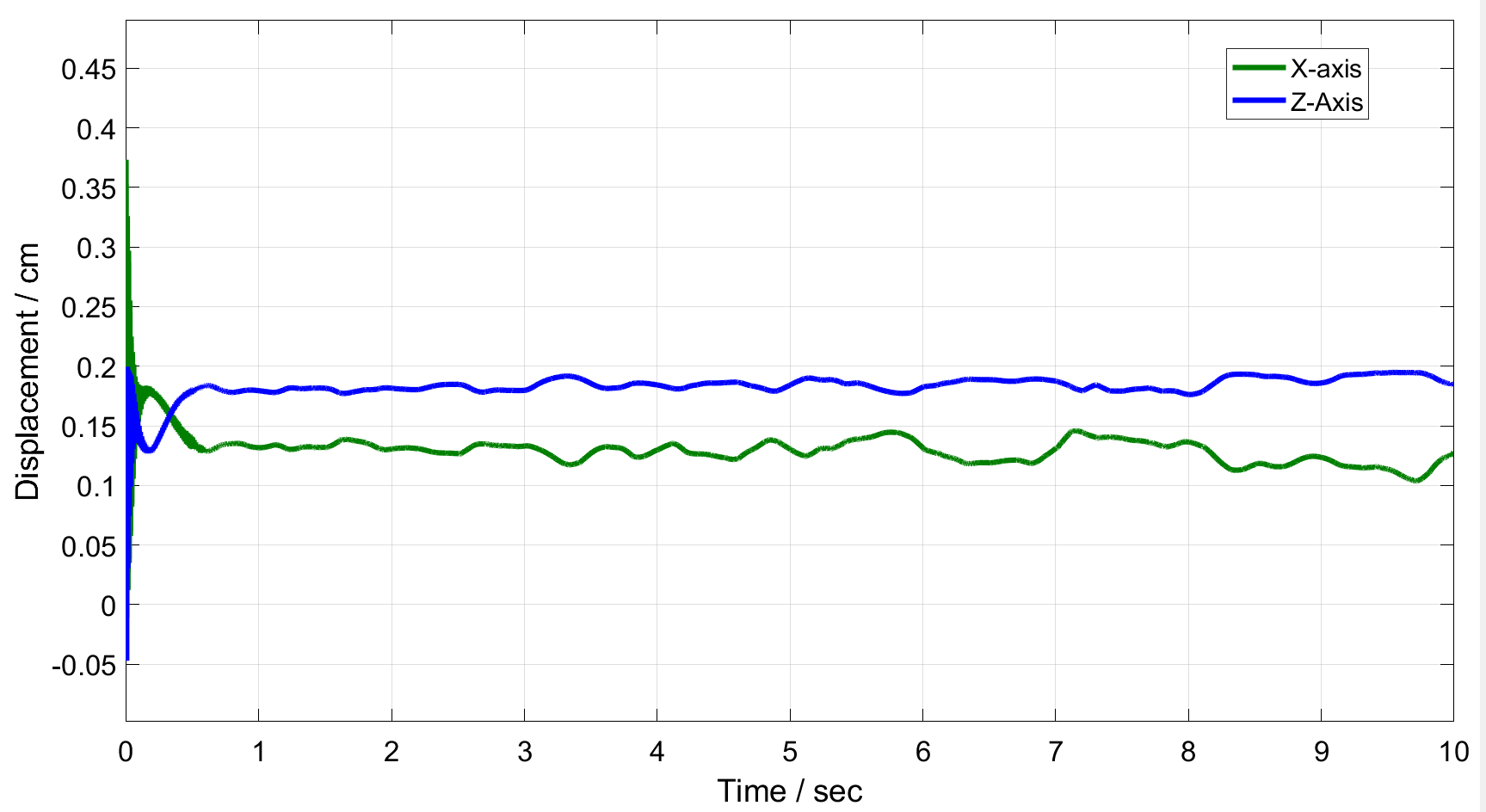}
}
      \caption{End-point tracking in XZ-axes for AMM system with a 2-DOF manipulator under disturbances.}
      \label{endpoint}
  \end{figure}

The AMM system with Dexterous Hexrotor and 2-DOF manipulator are modelled using Matlab Simscape toolbox. A PID controller is wrapped around this model along with the dynamics of the UAV and manipulator. Noise is introduced as the disturbance into the system to evaluate the performance of the AMM system in position tracking. The x \& z positions of the end-point are tracked and plotted in fig. \ref{endpoint}.

\section{CONCLUSIONS}

In this paper we have presented a macro-micro combination of AMM system to enable dexterous interactions with higher precision. The proposed fully-actuated UAV enables both the mobility and manipulability aspects of aerial mobile manipulation. A parallel manipulation as a part of AMM system, extends dexterity with increased precision. Operational space formulations are employed to decompose the joint \& rotor space into forces at the end-effector and motions in null-space. 

The AMM system to withstand internal and external disturbances caused by the motions of on-board manipulator, physical interactions, winds, etc. And hold the end-point with respect to the UAV frame despite of these disturbances. To support the claims, we have presented experimental results on the performance of the UAV in withstanding on-board manipulator motions and external winds. Additionally, we presented simulation results of end-point tracking for the AMM system with Dexterous Hexrotor and 2-DOF planar manipulator. This would further be extended to the 6-DOF manipulator to achieve higher precision.

\addtolength{\textheight}{-12cm}   




\section*{ACKNOWLEDGMENT}

This work was supported by the Dept. of Energy through the NSF Center for Robots and Sensor for the Human WellBeing (RoSe-HUB) and by the National Science Foundation under grants CNS-1450432, CNS-1138674 and IIS-1111568.


\end{document}